\documentclass[twoside,11pt]{article}
\usepackage{jair, rawfonts}

\usepackage{booktabs} 
\usepackage{tikz} 
\usepackage{amsmath,amssymb,amsthm,textcomp}
\usepackage{multicol}
\usepackage{stackengine}
\usepackage{amssymb}
\usepackage{pifont}
\usepackage{amsmath}
\usepackage{natbib}
\usepackage{mathtools}

\DeclareMathOperator*{\argmax}{arg\,max}

\newcommand{\norm}[1]{\left\lVert#1\right\rVert}
\newtheorem{definition}{Definition}[section]

\jairheading{1-22}{2024}{}{09/07}{}
\ShortHeadings{A Survey Analyzing Generalization in Deep Reinforcement Learning}
{Korkmaz}
\firstpageno{1}

\begin{document}

\title{A Survey Analyzing Generalization in \\ Deep Reinforcement Learning}

\author{\name Ezgi Korkmaz \\
       \addr University College London \\
       London, United Kingdom
       }

\maketitle

\begin{abstract}
Reinforcement learning research obtained significant success and attention with the utilization of deep neural networks to solve problems in high dimensional state or action spaces. While deep reinforcement learning policies are currently being deployed in many different fields from medical applications to large language models, there are still ongoing questions the field is trying to answer on the generalization capabilities of deep reinforcement learning policies. 
In this paper, we will formalize and analyze generalization in deep reinforcement learning. We will explain the fundamental reasons why deep reinforcement learning policies encounter overfitting problems that limit their generalization capabilities. Furthermore, we will categorize and explain the manifold solution approaches to increase generalization, and overcome overfitting in deep reinforcement learning policies. From exploration to adversarial analysis and from regularization to robustness our paper provides an analysis on a wide range of subfields within deep reinforcement learning with a broad scope and in-depth view. 
We believe our study can provide a compact guideline for the current advancements in deep reinforcement learning, and help to construct robust deep neural policies with higher generalization skills.
\end{abstract}

\section{Introduction}

The performance of reinforcement learning algorithms \citep{watkins89, sutton84,sutton88} has been boosted with the utilization of deep neural networks as function approximators \citep{mn15}. Currently, it is possible to learn deep reinforcement learning policies that can operate in large state and/or action space MDPs \citep{silver17, vinyals19}. This progress consequently resulted in building reasonable deep reinforcement learning policies that can play computer games with high dimensional state representations (e.g. Atari, StarCraft), solve complex robotics control tasks, design algorithms \citep{daniel23, fawzi22}, guide large language models \citep{openai23,gemini23}, and play some of the most complicated board games (e.g. Chess, Go) \citep{jul20nat}.
However, deep reinforcement learning algorithms also experience several problems caused by their overall limited generalization capabilities.
Some studies demonstrated these problems via adversarial perturbations introduced to the state observations of the policy \citep{huang17, kos17, korkmaz22, korkmaz2023icml}, several focused on exploring the fundamental issues with function approximation, estimation biases in the state-action value function \citep{trun93,hasselt10}, or with new architectural design ideas \citep{wang16}.
The fact that we are not able to completely explore the entire MDP for high dimensional state representation MDPs, even with deep neural networks as function approximators, is one of the root problems that limits generalization. On top of this, some portion of the problems are directly caused by the utilization of deep neural networks and thereby the intrinsic problems inherited from their utilization \citep{fellow15,szegedy13,korkmaz22, korkmazthesis24}.

In order to address open questions on generalization in deep reinforcement learning, there needs to be some commonly agreed standard of what is meant by generalization. Currently, different aspects of generalization are considered in various subfields either working on the fundamental questions regarding or the applications of deep reinforcement learning. We take the point of view in this paper that these various aspects can, and should, be described and studied in a unified way. In particular, we argue that the various approaches to generalization can be succinctly classified based on which part of the Markov Decision Process is expected to vary. We make this classification formal and unify how much current work on generalization in deep reinforcement learning fits clearly into the classification we introduce.
In this paper we will focus on generalization in deep reinforcement learning and the underlying causes of the limitations deep reinforcement learning research currently faces. In particular, we will try to answer the following questions:
\begin{itemize}
\item \textit{How can we formalize the concept of generalization in deep reinforcement learning?}
\item \textit{What is the role of exploration in overfitting for deep reinforcement learning?}
\item \textit{What are the causes of overestimation bias observed in state-action value functions?}
\item \textit{What has been done to overcome the overfitting problems that deep reinforcement learning algorithms have encountered so far, and to enable deep neural policies to generalize to non-stationary complex environments?}
\end{itemize}

To answer these questions we will go through research connecting several subfields in reinforcement learning on the problems and corresponding proposed solutions regarding generalization. In this paper we introduce a formal definition of generalization and categorization of the different methods used to both achieve and assess generalization, and use it to systematically summarize and consolidate the current body of research. 
We further describe the issue of value function overestimation, and the role of exploration in overfitting in reinforcement learning.
Furthermore, we explain new emerging research areas that can potentially target these questions in the long run including meta-reinforcement learning and lifelong learning.
The objective of the paper is to introduce a formal generalization definition and provide a compact overview and unification of the current advancements and limitations in the field.

\section{Preliminaries on Deep Reinforcement Learning}

The aim in deep reinforcement learning is to learn a policy via interacting with an environment in a Markov Decision Process (MDP) that maximize expected cumulative discounted rewards. An MDP is represented by a tuple $\mathcal{M} = (S,A,\mathcal{P},r,\rho_0, \gamma)$, where $S$ represents the state space, $A$ represents the action space, $r:S\times A \to \mathbb{R}$ is a reward function, $\mathcal{P}: S \times A \to \Delta(S)$ is a transition probability kernel, $\rho_0$ represents the initial state distribution, and $\gamma$ represents the discount factor.
The objective in reinforcement learning is to learn a policy $\pi:S \times A \to \mathbb{R}$ which maps states to probability distributions on actions in order to maximize the expected cumulative reward $R = \mathbb{E}\sum_{t=0}^{T-1}\gamma^t r(s_t,a_t)$ where $a_t \sim \pi(s_t, \cdot), s_{t+1} \sim \mathcal{P}(\cdot| s_t,a_t)$. The temporal difference updates achieves this objective by updating the value function $V(s)$ \citep{sutton84,sutton88}
\begin{equation}
\label{tdupdate}
V(s_t) \leftarrow V(s_t) + \alpha [r(s_{t+1},a) +\gamma V(s_{t+1}) - V(s_t)]
\end{equation}
While Equation \ref{tdupdate} represents the one-step temporal difference update, i.e. TD(0), it is further possible to consider multi-step TD which focuses on multi-step return, i.e. TD($\lambda$).
In $Q$-learning the goal is to learn the optimal state-action value function \citep{watkins89}
\begin{equation}
Q^*(s,a) = r(s,a) + \sum_{s' \in S} \mathcal{P}(s'|s,a) \max_{a' \in A} Q^*(s',a').
\end{equation}
This is achieved via iterative Bellman update \citep{bellman57,bellman59} which updates $Q(s_t,a_t)$ by 
\[
 Q(s_t,a_t) + \alpha [ \mathcal{R}_{t+1}+ \gamma \max_a Q(s_{t+1}, a) -Q(s_t,a_t) ].
\]
Thus, the optimal policy is determined by choosing the action $a^*(s) = \argmax_a Q(s,a)$ in state $s$. The optimal Bellman operator is \citep{bellman57}
\[
\mathcal{B} Q(s,a) \coloneqq  \mathbb{E} [r(s,a)] + \gamma \mathbb{E}_\mathcal{P} [\max_{a'} Q(s',a')]
\]
In high dimensional state space or action space MDPs the optimal policy is decided via a function-approximated state-action value function represented by a deep neural network.
The loss function in deep reinforcement learning is the quadratic difference between that target network and the current state-action value function.
\[
\mathcal{L}_i(\theta_i) = \mathbb{E}_{e \sim \mathcal{D}} [(r(s,a) +\gamma \max_{a'} Q(s',a',\theta_{\textrm{target}}) -Q(s,a,\theta_i))^2]
\]
where $\mathcal{D}$ is the experience replay buffer \citep{lin93} in which the experiences $e= \{ s_t,a_t,r_t,s_{t+1}\}$ sampled from $\mathcal{D} = \{ e_1, e_2, \dots e_N \}$. The loss function is optimized by taking the gradient with respect function approximation weights
\begin{align*}
\theta_{i+1} = \theta_{i} + \alpha (r(s_t,a_t, s_{t+1})
+ \gamma \mathcal{Q}(s_{t+1}, &\argmax_a \mathcal{Q}(s_{t+1},a;\theta^{\textrm{target}}_{i-1});\theta^{\textrm{target}}_{i-1}) \\
& \qquad\qquad \qquad  - \mathcal{Q}(s_t,a_t;\theta_i)) \nabla_{\theta_i} \mathcal{Q}(s_t,a_t;\theta_i).
\end{align*}
In a parallel line of algorithm families the policy itself is directly parametrized by $\pi_\theta$ \citep{sutton99}, and the gradient estimator used in learning is
\[
g = \mathbb{E}_t \big[ \nabla_\theta \log \pi_\theta(s_t,a_t) (Q(s_t,a_t) - \max_a Q(s_t,a) ) \big]
\]
where $Q(s_t,a_t)$ refers to the state-action value function at time step $t$.
The algorithms that focus on directly parameterizing the policy try to solve the following optimization problem \citep{schulman15}.
\begin{align*}
  \max_\theta \mathbb{E}_{s \sim \rho_{\theta_{\textrm{old}}}; a\sim \pi_{\theta_{\textrm{old}}}(s,\cdot)} \left[ \dfrac{\pi_\theta(s,a)}{\pi_{\theta_{\textrm{old}}}(s,a)} Q_{\theta_{\textrm{old}}}(s,a) \right]
  \:\: \textrm{subject to} \:\: & \mathbb{E}_{s \sim \rho_{\theta_{\textrm{old}}}} \mathcal{D}_{KL}(\pi_\theta(s,\cdot) || \pi_{\textrm{old}}(s,\cdot)) \leq \delta
\end{align*}

\section{How to Achieve Generalization?}
\label{def}

\subsection{Generic Reinforcement Learning Algorithm}
To be able to understand and analyze the connection between different approaches to achieve generalization first we will provide a clear definition intended to capture the behavior of a generic reinforcement learning algorithm.
\begin{definition}[\emph{Generic reinforcement learning algorithm}]
  A reinforcement learning training algorithm $\mathcal{A}$ learns a policy $\pi$ by interacting with an MDP $\mathcal{M}$. We divide up the execution of $\mathcal{A}$ into discrete time steps as follows. At each time $t$, the algorithm has a current policy $\pi_t$, observes a state $s_t$, takes an action $a_t \sim \pi_t(s_t, \cdot)$, and observes a transition to state $s'_t\sim \mathcal{P}(\cdot\mid s_t, a_t)$ with corresponding reward $r_t = r(s_t,a_t,s'_t)$.
  We define the history of algorithm $\mathcal{A}$ in MDP $\mathcal{M}$ to be the sequence $H_t = (\pi_0,s_0,a_0,s'_0,r_0),\dots (\pi_t,s_t,a_t,s'_t,r_t)$ of all the transitions observed by the algorithm so far.
  We require that the policy $\pi_t$ and state $s_t$ at time $t$ are a function only of $H_{t-1}$, i.e the transitions observed so far by $\mathcal{A}$.
  At time $t=T$, the algorithm stops and outputs the policy $\pi=\pi_T$. We use the notation $\mathbb{A}$ to denote the set of reinforcement learning training algorithms and $\Pi$ to denote the set of policies $\pi$ in an MDP $\mathcal{M}$.
\end{definition}
Intuitively, a reinforcement learning algorithm has a current policy $\pi_t$, performs a sequence of queries $(s_t,a_t)$ to the MDP, and observes the resulting state transitions and rewards. In order to be as generic as possible, the definition makes no assumptions about how the algorithm chooses the sequence of queries, other than that $a_t \sim \pi_t(s_t, \cdot)$. Notably, if taking action $a_t$ in state $s_t$ leads to a transition to state $s'_t$, there is no requirement that $s_{t+1} = s'_t$. Indeed, the only assumption is that $s_{t+1}$ and $\pi_{t+1}$ may depend only on $H_t$, the history of transitions observed so far.
This allows the definition to capture deep reinforcement learning algorithms, which may choose to query states and actions in a complex way as a function of previously observed state transitions.

\subsection{Base Generalization in Deep Reinforcement Learning }

We next introduce a basic metric capturing how well an algorithm generalizes given a fixed amount of interaction with a given MDP.
\begin{definition}[\emph{Base generalization}]
  \label{def:basegeneralization}
  Given an MDP $\mathcal{M} = (S,A,P,r,\rho_0, \gamma)$, let $\pi_T$ and $\hat{\pi}_T$ be policies output by training algorithms taking $T$ steps.
  The base generalization $\mathcal{G}^{\textrm{base}}$ is the difference between the expected discounted cumulative rewards obtained by policy $\pi_T$ and $\hat{\pi}_T$ in $\mathcal{M}$.
  \begin{align*}
  \mathcal{G}^{\textrm{base}}(\pi_T, \hat{\pi}_T) = 
  \mathbb{E}_{a_t \sim \pi_T(s_t,\cdot)} &\left[\sum_{t=0}^{\infty}\gamma^t r(s_t,  a_t, s_{t+1})\right] \\
  & \qquad \quad - \mathbb{E}_{\hat{a}_t \sim \hat{\pi}_T(\hat{s}_t,\cdot)} \left[\sum_{t=0}^{\infty}\gamma^t r(\hat{s}_t,  \hat{a}_t, \hat{s}_{t+1})\right] 
  \end{align*}
\end{definition}
The base generalization definition captures how well an algorithm can generalize to unseen states and transitions, given only access to $T$ interactions with the MDP $\mathcal{M}$. Hence, in base generalization the role of exploration is exceedingly dominant and this will be further explained in Section \ref{exploration}.

\subsection{Algorithmic Generalization}

Based on the definition of generic reinforcement learning algorithm, we will now further define the different approaches proposed to achieve generalization.
At a high level, the approaches we will discuss will be divided into two classes: 
\vskip 0.1in
 \textbf{I.} Techniques that solely modify the training algorithm, \\
\indent \textbf{II.} Techniques that directly modify the MDP (i.e. learning environment, training data) that forms 
the interactions of the training algorithm with the learning environment. \\
\vskip 0.1in
Our first definition formalizes the techniques that solely modify the training algorithm.
\begin{definition}[\emph{Algorithmic generalization}]
  \label{def:algorithmicgeneralization}
    Let $\mathcal{A}$ be a training algorithm that takes an MDP as input and outputs a policy.
    Given an MDP $\mathcal{M} = (S,A,P,r,\rho_0, \gamma)$, an \emph{algorithmic} generalization method $\mathcal{G}_\mathbb{A}$ is given by a function $F: \mathbb{A} \to \mathbb{A}$ that runs the algorithm $F(\mathcal{A})$ in the MDP $\mathcal{M}$.
\end{definition}
Algorithmic generalization captures modifications to the training algorithm itself that can range from the choice of optimization methods or regularizers, to update rules for the policy.

\subsection{Generalization Through Rewards}

\begin{definition}[\emph{Rewards transforming generalization}]
\label{def:rewardsperturbing}
Let $\mathcal{A}$ be a training algorithm that takes as input an MDP and outputs a policy.
Given an MDP $\mathcal{M} = (S,A,P,r,\rho_0, \gamma)$, a \emph{rewards transforming} generalization method $\mathcal{G}_R$ is given by a sequence of functions $F_t:(\Pi \times S\times A\times S\times \mathbb{R})^t \times\mathbb{R} \to \mathbb{R}$. The method attempts to achieve generalization by running $\mathcal{A}$ on MDP $\mathcal{M}$, but modifying the rewards at each time $t$ to be $\hat{r}_t(s_t,a_t,s'_t) = F_{t-1}(H_{t-1},r_t)$, where $H_{t-1}$ is the history of algorithm $\mathcal{A}$ when running with the transformed rewards.
\end{definition}

In particular, a method under the rewards transforming generalization category runs the original algorithm to train the policy, but modifies the observed rewards. The instances of these techniques will be mentioned and explained in Section \ref{direct-gen}, i.e. direct function regularization, in Section \ref{exploration}, i.e. the role of exploration in overfitting, and in Section \ref{transfer}, i.e. transfer in reinforcement learning.

\subsection{Generalization Through Observations}

Following the definition of reward transforming generalization we define state transforming generalization which is one of the canonical approaches for achieving generalization in deep reinforcement learning. The instances of generalization through observations will be categorized and explained in detail in Section \ref{data-aug}, i.e. data augmentation, and Section \ref{adv}, i.e. the adversarial perspective for deep neural policy generalization.
\begin{definition}[\emph{State transforming generalization}]
\label{def:stateperturbing}
  Let $\mathcal{A}$ be a training algorithm that takes as input an MDP and outputs a policy.
  Given an MDP $\mathcal{M} = (S,A,P,r,\rho_0, \gamma)$, a \emph{state transforming} generalization method $\mathcal{G}_S$ is given by a sequence of functions $F_t:(\Pi \times S\times A\times S\times \mathbb{R})^t\times S\to S$.
  The method attempts to achieve generalization by running $\mathcal{A}$ on MDP $\mathcal{M}$, but modifying the state chosen at time $t$ to be $\hat{s}_t = F_{t-1}(H_{t-1},s_t)$, where $H_{t-1}$ is the history of algorithm $\mathcal{A}$ when running with the transformed states.
\end{definition}

\subsection{Generalization Through Environment Dynamics}
Another category of algorithms that tries to achieve generalization in deep reinforcement learning focuses on achieving this objective through environment dynamics transformation. The methods focusing on generalization through environment dynamics will be referred to and explained in Section \ref{direct-gen}, i.e. direct function regularization.

\begin{definition}[\emph{Transition probability transforming generalization}]
\label{def:transitionperturbing}
  Let $\mathcal{A}$ be a training algorithm that takes as input an MDP and outputs a policy.
  Given an MDP $\mathcal{M} = (S,A,P,r,\rho_0, \gamma)$, a \emph{transition probability transforming} generalization method $\mathcal{G}_\mathcal{P}$ is given by a sequence of functions $F_t:(\Pi \times S\times A\times S\times \mathbb{R})^t\times (S\times A\times S)\to \mathbb{R}$.
  The method attempts to achieve generalization by running $\mathcal{A}$ on MDP $\mathcal{M}$, but modifying the transition probabilities at time $t$ to be $\hat{P}(s_t,a_t,s'_t) = F_{t-1}(H_{t-1},s_t,a_t,s'_t)$, where $H_{t-1}$ is the history of algorithm $\mathcal{A}$ when running with the transformed transition probabilities.
\end{definition}

\subsection{Generalization Through Policy}
The last type of generalization method we define is based on directly modifying the current policy used by the algorithm to select actions at each time step. We will explain the instances of the techniques that focus on generalization through policy in Section \ref{exploration}, i.e. the role of exploration in overfitting, and Section \ref{metarlsec}, i.e. meta reinforcement learning and meta gradients.
\begin{definition}[\emph{Policy transforming generalization}]
\label{def:policyperturbing}
Let $\mathcal{A}$ be a training algorithm that takes as input an MDP and outputs a policy.
Given an MDP $\mathcal{M} = (S,A,\mathcal{P},r,\rho_0, \gamma)$, a \emph{policy transforming} generalization method $\mathcal{G}_\pi$ is given by a sequence of functions $F_t:(\Pi \times S\times A\times S\times \mathbb{R})^t \times S \times \Delta(A) \to \Delta(A) $.
The method attempts to achieve generalization by running $\mathcal{A}$ on MDP $\mathcal{M}$, but modifying the current policy by which $\mathcal{A}$ chooses the action at time $t$ to be $\hat{\pi_t}(s_t,\cdot) = F_{t-1}(H_{t-1},s_t,\pi_t(s_t,\cdot))$, where $H_{t-1}$ is the history of algorithm $\mathcal{A}$ when running with the transformed policy.
\end{definition}

\subsection{Assessing Generalization}

All the definitions so far categorize methods to modify either training algorithms and/or the MDP, i.e. learning environment, training data, in order to achieve generalization. However, many such methods for modifying training algorithms have a corresponding method which can be used to assess the generalization capabilities of a trained policy. Our final definition captures this correspondence.

\begin{definition}[\emph{Generalization testing}]
\label{def:gentesting}
Let $\hat{\pi}$ be a trained policy for an MDP $\mathcal{M}$. Let $F_t$ be a sequence of functions corresponding to a generalization method from one of the previous definitions. The \emph{generalization testing} method of $F_t$ is given by executing the policy $\hat{\pi}$ in $\mathcal{M}$, but in each time step applying the modification $F_t$ where the history $H_t$ is given by the transitions executed by $\hat{\pi}$ so far. When both a generalization method and a generalization testing method are used concurrently, we will use subscripts to denote the generalization method and superscripts to denote the testing method. For instance, $\mathcal{G}_S^{\pi}$ corresponds to training with a state transforming method, and testing with a policy transforming method.
\end{definition}

\section{Roots of Overestimation in Deep Reinforcement Learning}

Many reinforcement learning algorithms compute estimates for the state-action values in an MDP. Because these estimates are usually based on a stochastic interaction with the MDP, computing accurate estimates that correctly generalize to further interactions is one of the most fundamental tasks in reinforcement learning. A major challenge in this area has been the tendency of many classes of reinforcement learning algorithms to consistently overestimate state-action values.
Initially the overestimation bias for $Q$-learning is discussed and theoretically justified by \cite{trun93} as a biproduct of using function approximators for state-action value estimates.
In particular, \cite{trun93} proves that if the reinforcement learning policy overestimates the state-action values by $\gamma c$ during learning then the Q-learning algorithm will fail to learn optimal policy if $\gamma> \dfrac{1}{1+c}$.

Following this initial discussion it has been shown that several parts of the deep reinforcement learning process can cause overestimation bias. Learning overestimated state-action values can be caused by statistical bias of utilizing a single max operator \citep{hasselt10}, coupling between value function and the optimal policy \citep{fergus21, cobbe21}, or caused by the accumulated function approximation error \citep{moore94}.

Several methods have been proposed to target overestimation bias for value iteration algorithms. 
In particular, \cite{hasselt10} demonstrated that the expectation of a maximum of a random variable is not equal to maximum of the expectation of a random variable.
\[
\mathbb{E} [ \max_i [X_i]] \neq  \max_i  [\mathbb{E}  [X_i] ] \:\: \textrm{where} \:\: X = \{X_1,X_2,\dots,X_N\}
\]
This clear distinction shows that simple Q-learning is a biased estimator, and to solve this overestimation bias introduced by the max operator \cite{hasselt10} proposed to utilize a double estimator for the state-action value estimates. In particular, the double estimator for double Q-learning works as follows
\[
Q^{\textrm{I}}(s,a) \leftarrow Q^{\textrm{I}}(s,a) + \alpha(s,a) (r(s,a) + \gamma Q^{\textrm{II}}(s', \max_a Q^{\textrm{I}}(s',a)) -  Q^{\textrm{I}}(s',a))
\]
and 
\[
Q^{\textrm{II}}(s,a) \leftarrow Q^{\textrm{II}}(s,a) + \alpha(s,a) (r(s,a) + \gamma Q^{\textrm{I}}(s', \max_a Q^{\textrm{II}}(s',a)) -  Q^{\textrm{II}}(s',a)).
\]
Later, the authors also created a version of this algorithm that can solve high dimensional state space problems \citep{hado16}. Some of the work on this line of research targeting overestimation bias for value iteration algorithms is based on simply averaging the state-action values with previously learned state-action value estimates during training time \citep{oron17}.
While overestimation bias was demonstrated to be a problem and discussed over a long period of time \citep{trun93,hasselt10}, recent studies also further demonstrated that actor critic algorithms also suffer from this issue \citep{meger18}.

\section{The Role of Exploration in Overfitting}
\label{exploration}
The fundamental trade-off of exploration vs exploitation is the dilemma that the agent can try to take actions to move towards more unexplored states by sacrificing the current immediate rewards. While there is a significant body of studies on provably efficient exploration strategies the results from these studies do not necessarily directly transfer to the high dimensional state or action MDPs. The most prominent indication of this is that, even though it is possible to use deep neural networks as function approximators for large state spaces, the agent will simply not be able to explore the full state space. The fact that the agent is able to only explore a portion of the state space simply creates a bias in the learnt value function \citep{baird95}.

\begin{table}[t]
\caption{Environment and algorithm details for different exploration strategies for generalization.}
\label{percentageshift}
\centering
\scalebox{0.9}{
\begin{tabular}{lccccccr}
\toprule
Citation & Method & Learning Environment & Algorithm  \\
\midrule
\cite{mn15}       &  $\epsilon$-greedy  & Arcade Learning Environment    &    DQN \\
\cite{bell16nips} & Count-based         & Arcade Learning Environment    & A3C and DQN \\
\cite{osband16}   & RLSVI               & Tetris      & Tabular $Q$ \\
\cite{osband16nips} & Bootstrapped DQN  & Arcade Learning Environment & DQN \\
\cite{hout17}      & VIME & DeepMind Control Suite & TRPO \\
\cite{fortun18}    & NoisyNet   & Arcade Learning Environment                 & A3C and DQN \\
\cite{lee21}       & SUNRISE    & DCS\footnotemark \& ALE                     & SAC \& RDQN \\
\cite{mahankali24} & Random Latent   & Arcade Learning Environment & PPO \\
\bottomrule
\end{tabular}
}
\end{table}
\footnotetext{DeepMind Control Suite}

In this section, we will go through several exploration strategies in deep reinforcement learning and how they affect policy overfitting. A quite simple version of this is based on adding noise in action selection during training e.g. $\epsilon$-greedy exploration. Note that this is an example of a policy transforming generalization method $\mathcal{G}_\pi$ in Definition \ref{def:policyperturbing} in Section \ref{def}.
While $\epsilon$-greedy exploration is widely used in deep reinforcement learning \citep{wang16,jess20,hado23}, it has also been proven that to explore the state space these algorithms may take exponentially long \citep{kakade03}.
Several others focused on randomizing different components of the reinforcement learning training algorithms. In particular, \cite{osband16} proposes the randomized least squared value iteration algorithm to explore more efficiently in order to increase generalization in reinforcement learning for linearly parametrized value functions. This is achieved by simply adding Gaussian noise as a function of state visitation frequencies to the training dataset. Later, the authors also propose the bootstrapped DQN algorithm (i.e. adding temporally correlated noise) to increase generalization with non-linear function approximation \cite{osband16nips}.
Recently, \cite{mahankali24} proposed to randomize the reward function to enhance exploration in high dimensional observation MDPs where policy gradient algorithms are used to explore. This study is also a clear example of the generalization through rewards as has been explained in Definition \ref{def:rewardsperturbing} in Section \ref{def}.

\cite{hout17} proposed an exploration technique centered around maximizing the information gain on the agent's belief of the environment dynamics. In practice, the authors use Bayesian neural networks for effectively exploring high dimensional action space MDPs.
Following this line of work on increasing efficiency during exploration \cite{fortun18} proposes to add parametric noise to the  deep reinforcement learning policy weights in high dimensional state MDPs.
While several methods focused on ensemble state-action value function learning \citep{osband16nips}, \cite{lee21} proposed reweighting target Q-values from an ensemble of policies (i.e. weighted Bellman backups) combined with highest upper-confidence bound action selection.
Another line of research in exploration strategies focused on \textit{count-based methods} that use the direct count of state visitations. In this line of work, \cite{bell16nips} tried to lay out the relationship between count based methods and intrinsic motivation, and used count-based methods for high dimensional state MDPs (i.e. Arcade Learning Environment). Yet it is worthwhile to note that most of the current deep reinforcement learning algorithms use very simple exploration techniques such as $\epsilon$-greedy which is based on taking the action maximizing the state-action value function with probability $1-\epsilon$ and taking a random action with probability $\epsilon$ \citep{mn15,hado16,wang16,jess20,hado23}.

It is possible to argue that the fact that the deep reinforcement learning policy obtained a higher score with the same number of samples by a particular type of training method $\mathcal{A}$ compared to method $\mathcal{B}$ is by itself evidence that the technique $\mathcal{A}$ leads to more generalized policies. Even though the agent is trained and tested in the same environment, the explored states during training time are not exactly the same states visited during test time. The fact that the policy trained with technique $\mathcal{A}$ obtains a higher score at the end of an episode is sole evidence that the agent trained with $\mathcal{A}$ was able to visit further states in the MDP and thus succeed in them. Yet, throughout the paper we will discuss different notions of generalization investigated in different subfields of reinforcement learning research.
While exploration vs exploitation stands out as one of the main problems in reinforcement learning policy performance most of the work conducted in this section focuses on achieving higher score in hard-exploration games (i.e. Montezuma's Revenge) rather than aiming for a generally higher score for each game overall across a given benchmark. Thus, it is possible that the majority of work focusing on exploration so far might not be able to obtain policies that perform as well as those in the studies described in Section \ref{reg} across a given benchmark.

\section{Regularization}
\label{reg}
In this section we will focus on different regularization techniques employed to increase generalization in deep reinforcement learning policies. We will go through these works by categorizing each of them under data augmentation, adversarial training, and direct function regularization. Under each category we will connect these different lines of approach to increase generalization in deep reinforcement learning to the settings we defined in Section \ref{def}.

\begin{table}[t]
  \caption{Environment and algorithm details for data augmentation techniques for state observation generalization. All of the studies in this section focus on state transformation methods $\mathcal{G}_S$ defined in Section \ref{def}.}
  \label{percentageshift}
  \centering
  \scalebox{0.92}{
  \begin{tabular}{lccccccr}
  \toprule
  Citation &  Method & Environment & Algorithm  \\
  \midrule
  \cite{yarats21}      & DrQ                  & DCS, Arcade Learning Environment & DQN \\
  \cite{laskin20icml}  & CuRL                 & DCS, Arcade Learning Environment & SAC and DQN \\
  \cite{laskin20}      & RAD                  & DeepMind Control Suite, ProcGen & SAC and PPO \\
\cite{korkmaz2023aaai} & Semantic Changes     & Arcade Learning Environment     & DDQN \& A3C \\
  \cite{feng20}        & Mixreg               & ProcGen & DQN and PPO \\
  \bottomrule
  \end{tabular}
  }
  \end{table}

\subsection{Data Augmentation}
\label{data-aug}
Several studies focus on diversifying the observations of the deep reinforcement learning policy to increase generalization capabilities. A line of research in this regard focused on simply employing versions of data augmentation techniques \citep{laskin20,laskin20icml,yarats21} for high dimensional state representation environments. In particular, these studies involve simple techniques such as cropping, rotating or shifting the state observations during training time. While this line of work got considerable attention, a quite recent study \cite{bell21} demonstrated that when the number of random seeds is increased to one hundred the relative performance achieved and reported in the original papers of \citep{laskin20icml,yarats21} on data augmentation training in deep reinforcement learning decreases to a level that might be significant to mention.

While some of the work on this line of research simply focuses on using a set of data augmentation methods \citep{laskin20, laskin20icml, yarats21}, other work focuses on proposing new environments to train in \citep{cobbe19}. The studies on designing new environments to train deep reinforcement learning policies basically aim to provide high variation in the observed environment such as changing background colors and changing object shapes in ways that are meaningful in the game, in order to increase test time generalization.
In the line of robustness and test time performance, a more recent work that is also mentioned in Section \ref{adversarialsurvey} demonstrated that imperceptible semantically meaningful data augmentations can cause significant damage on the policy performance and certified robust deep reinforcement learning policies are more vulnerable to these imperceptible augmentations \citep{korkmaz2021natural,korkmaz2023aaai}.

Within this category some work focuses on producing more observations by simply blending in (e.g. creating a mixture state from multiple different observations) several observations to increase generalization \citep{feng20}. While most of the studies trying to increase generalization by data augmentation techniques are primarily conducted in the DeepMind Control Suite or the Arcade Learning Environment (ALE) \citep{bell13}, some small fraction of these studies \citep{feng20} are conducted in relatively recently designed training environments like ProcGen \citep{cobbe19}. 
In the line of research proposing learning environments \cite{dennis20} proposed unsupervised environment design by changing the environment parameters to asses generalization for maze structured environments by minimax training where the "adversary" creating an environment for the policy to solve a task with goal and obstacles as an underspecified parameter.
\cite{cobbe19icml} focuses on decoupling the training and testing set for reinforcement learning via simply proposing a new game environment CoinRun.

\begin{table*}[t]
\caption{Environment and algorithm details for different direct function regularization strategies for trying to overcome overfitting problems in reinforcement learning. Note that most of the methods based on direct function regularization are a form of algorithmic generalization $\mathcal{G}_\mathbb{A}$ to overcome overfitting as described in Section \ref{def}.}
\label{percentageshift}
\centering
\hskip -0.2in
\scalebox{0.87}{
\begin{tabular}{lccccccr}
\toprule
Citation & Proposed Method & Learning Environment  \\
\midrule
\cite{maxim19} & SNI and IBAC & GridWorld and CoinRun \\
\cite{viel20} & Munchausen RL  & Arcade Learning Environment  \\
\cite{lee20} & Network Randomization & 2D CoinRun and 3D DeepMind Lab \\
\cite{amit20} & Discount Regularization & GridWorld and MuJoCo\footnotemark  \\
\cite{agarwal21} & PSM   & DDMC and Rectangle Game\footnotemark   \\
\cite{liu21} & BN and dropout and $L_2/L_1$  & MuJoCo   \\
\bottomrule
\end{tabular}
}
\end{table*}

\addtocounter{footnote}{-1}

\footnotetext{Low dimensional setting of MuJoCo is used for this study \citep{todorov2012mujoco}.}
\addtocounter{footnote}{1}

\footnotetext{Rectangle game is a simple video game with only two actions, ”Right” and ”Jump”. The game has black background and two rectangles where the goal of the game is to avoid white obstacles and reach to the right side of the screen. \cite{agarwal21} is the only paper we encountered experimenting with this particular game.}

\subsection{Direct Function Regularization}
\label{direct-gen}
While some of the work we have discussed so far focuses on regularizing the data (i.e. state observations) as in Section \ref{data-aug}, some focuses on directly regularizing the function learned with the intention of simulating techniques from deep neural network regularization like batch normalization and dropout \citep{maxim19}. While some studies have attempted to simulate these known techniques in reinforcement learning, some focus on directly applying them to overcome overfitting. In this line of research, \cite{liu21} proposes to use known techniques from deep neural network regularization to apply in continuous control deep reinforcement learning training. In particular, these techniques are batch normalization (BN) \citep{iof15}, weight clipping, dropout, entropy and $L_2/L_1$ weight regularization. All these methods fall under the algorithmic generalization category $\mathcal{G}_\mathbb{A}$ as described in Section\ref{def}.

\cite{lee20} proposes to utilize a random network to essentially achieve a version of randomization in the input observations to increase generalization skills of deep reinforcement learning policies, and tests the proposal in the 2D CoinRun game proposed by \cite{cobbe19icml} and 3D DeepMind Lab \citep{deepmindlab}. In particular, the authors essentially introduce a random convolutional layer to achieve this objective.
This study is an example of an algorithmic generalization method $\mathcal{G}_\mathbb{A}$ described in Definition \ref{def:algorithmicgeneralization} when the single layer random network is not placed at the first layer of the deep neural network.
However, when this single layer random network is placed at the first layer of the neural network, this method is essentially just introducing some noise to the state observations of the policy, thus this is an example of state transforming generalization. 
When this single random layer is placed other than first, the method is no longer a state transforming generalization method because the states are not modified before they have been observed by the algorithm, but rather implicitly changed due to a random convolutional layer added in the architecture. We will further provide clear instances of the state transformation generalization also in Section \ref{adversarialsurvey} when the worst-case perturbation methods to target generalization in reinforcement learning policies are explained.

\begin{table*}[t]
\caption{Algorithm details for different direct function regularization strategies for trying to overcome overfitting problems in reinforcement learning. Note that most of the methods based on direct function regularization are a form of algorithmic generalization $\mathcal{G}_\mathbb{A}$ to overcome overfitting as described in Section \ref{def}.}
\label{directfunction}
\centering
\hskip -0.2in
\scalebox{0.87}{
\begin{tabular}{lccccccr}
\toprule
Citation & Proposed Method  & Reinforcement Learning Algorithm  \\
\midrule
\cite{maxim19} & SNI and IBAC & Proximal Policy Optimization (PPO) \\
\cite{viel20} & Munchausen RL  & DQN and IQN \\
\cite{lee20} & Network Randomization& Proximal Policy Optimization (PPO) \\
\cite{amit20} & Discount Regularization  & Twin Delayed DDPG (TD3) \\
\cite{agarwal21} & PSM     & Data Regularized-Q (DrQ) \\
\cite{liu21} & BN and dropout and $L_2/L_1$  & PPO, TRPO, SAC, A2C  \\
\bottomrule
\end{tabular}
}
\end{table*}

Some work employs contrastive representation learning to learn deep reinforcement learning policies from state observations that are close to each other \citep{agarwal21}.
This study leverage the temporal aspect of reinforcement learning and propose a policy similarity metric. The main goal of the paper is to lay out the sequential structure and utilize representation learning to learn generalizable abstractions from state representations.
One drawback of this study is that most of the experimental study is conducted in a non-baseline environment (i.e. Rectangle game and Distracting DM Control Suite). 
\cite{malik21} studies query complexity of reinforcement learning policies that can generalize to multiple environments. The authors of this study focus on an example of the transition probability transformation setting $\mathcal{G}_{\mathcal{P}}$ in Definition \ref{def:transitionperturbing}, and the reward function transformation setting $\mathcal{G}_R$ in Definition \ref{def:rewardsperturbing}.

Another line of study in direct function generalization investigates the relationship between reduced discount factor and adding an $\ell_2$-regularization term to the loss function, i.e. weight decay \citep{amit20}. The authors in this work demonstrate the explicit connection between reducing the discount factor and adding an $\ell_2$-regularizer to the value function for temporal difference learning. In particular, this study demonstrates that adding an $\ell_2$-regularization term to the loss function is equal to training with a lower discount term, which the authors refer to as \textit{discount regularization}. The results of this study however are based on experiments from tabular reinforcement learning, and the low dimensional setting of the MuJoCo environment \citep{todorov2012mujoco}.
This study is also another clear example of algorithmic generalization $\mathcal{G}_{\mathbb{A}}$ as described in Definition \ref{def:algorithmicgeneralization}.

On the reward transformation for generalization setting $\mathcal{G}_R$ defined in Definition \ref{def:rewardsperturbing}, \cite{viel20} adds the scaled log policy to the current rewards.
To overcome overfitting some work tries to learn explicit or implicit similarity between the states to obtain a reasonable policy \citep{le21}. In particular, the authors in this work try to unify the state space representations by providing a taxonomy of metrics in reinforcement learning.
Several studies proposed different ways to include Kullback-Leibler divergence between the current policy and the pre-updated policy to add as a regularization term in the reinforcement learning objective \citep{schulman15}. Recently, some studies argued that utilizing Kullback-Leibler regularization implicitly averages the state-action value estimates \citep{nino20}.

\begin{table}[t]
\caption{Environment and algorithm details for adversarial policy regularization and attack techniques in deep reinforcement learning. Note that most of the methods based on adversarial approaches are a form of generalization assessment through state observations $\mathcal{G}^S$ as described in Definition \ref{def:gentesting}, and some falls under the generalization through environment dynamics $\mathcal{G}_\mathcal{P}$ as described in Definition \ref{def:transitionperturbing}.}
\label{adversarial}
\centering
\scalebox{0.82}{
\begin{tabular}{lccccccr}
\toprule
Citation           &  Method & Environment & Algorithm  \\
\midrule
\cite{huang17}         & Fast Gradient Sign (FGSM)                          & ALE & DQN, TRPO, A3C \\
\cite{kos17}           & Fast Gradient Sign (FGSM)                          & ALE & DQN \& IQN \\
\cite{korkmaz22}       & Adversarial Framework                              & ALE  & DDQN \& A3C \\
\cite{lin17}           & Timing                                             & ALE  & A3C \& DQN \\
\cite{pinto17}         & Zero-sum game                                      & MuJoCo & RARL \\
\cite{glaeve19}        & Adversarial Policies                               & MuJoCo & PPO \\
\cite{korkmaz2023aaai} & Natural Attacks                                    & ALE      & DDQN \& A3C \\
\cite{huan20}          & State Adversarial-DQN                              & ALE and $L_{\textrm{M}}$\footnotemark & DDQN \& PPO \\
\cite{ezgikorkmazicml24}& Diagnostic Adversarial Volatility                  & ALE       & DDQN \\
\bottomrule
\end{tabular}
}
\end{table}

\footnotetext{Low dimensional state MuJoCo refers to the setting of MuJoCo where the state dimensions are not represented by pixels and dimensions of the state observations range from 11 to 117.}

\subsection{The Adversarial Perspective for Deep Neural Policy Generalization}
\label{adversarialsurvey}

One of the ways to regularize the state observations is based on considering worst-case perturbations added to state observations (i.e. adversarial perturbations). This line of work starts with introducing perturbations produced by the fast gradient sign method proposed by \cite{fellow15} into deep reinforcement learning observations at test time \citep{huang17, kos17}, and compares the generalization capabilities of the trained deep reinforcement learning policies in the presence worst-case perturbations and Gaussian noise. These gradient based adversarial methods are based on taking the gradient of the cost function used to train the policy with respect to the state observation.
\[
\mathnormal{\displaystyle s_{\textrm{adv}} = s + \epsilon \cdot \frac{\nabla_{x}J(\displaystyle s,Q(s,a))}{\norm{\nabla_{s}J(s,Q(s,a))}_p},}
\]
Several other techniques have been proposed on the optimization line of the adversarial alteration of state observations.
In this line of work, \cite{korkmaz20} suggested a Nesterov momentum-based method to produce adversarial perturbations for deep reinforcement learning policies.
\begin{align*}
  v_{t+1} = \mu \cdot v_t &+ \dfrac{\nabla_{s_\textrm{adv}}J(s_{\textrm{adv}}^t +  \mu \cdot v_t ,a)}{\lVert\nabla_{s_\textrm{adv}}J(s_{\textrm{adv}}^t +  \mu \cdot v_t ,a)\rVert_1} \\
   s_\textrm{adv}^{t+1} &= s_\textrm{adv}^{t} + \alpha \cdot  \dfrac{v_{t+1}}{\lVert v_{t+1}\rVert_2} \\
  \end{align*}
Here $J(s_{\textrm{adv}}, a)$ is based on the cost function used to train the policy, $s_{\textrm{adv}}$ represents the adversarial state observation, and $\mu$ is the momentum acceleration parameter. 
While a line of studies focused on optimization aspects of the adversarial perturbations, some studies demonstrated further the hidden linearity of deep reinforcement learning policies by revealing how these policies learn shared adversarial features across states, MDPs and across algorithms \citep{korkmaz22}.

In this work the authors investigate the root causes of this problem, and demonstrate that policy high-sensitivity directions and the perceptual similarity of the state observations are uncorrelated. Furthermore, the study demonstrates that the current state-of-the-art adversarial training techniques also learn similar high-sensitivity directions as the vanilla trained deep reinforcement learning policies.\footnote{From the security point of view, this adversarial framework is under the category of black-box adversarial attacks for which this is the first study that demonstrated that deep reinforcement learning policies are vulnerable to black-box adversarial attacks \citep{korkmaz22}. Furthermore, note that black-box adversarial perturbations are more generalizable global perturbations that can affect many different policies.}
More recently, a line of work proposed theoretically founded algorithms to understand the temporal and spatial correlation of deep reinforcement learning decision making and what affects this decision making process \citep{ezgikorkmazicml24}. In particular, this study identifies what precisely affects and contributes to the decision making process of deep reinforcement learning policies from distributional shift to worst-case perturbations (i.e. adversarial), from algorithmic differences to architectural changes.

While several studies focused on improving computation techniques to optimize optimal perturbations, a line of research focused on making deep neural policies resilient to these perturbations.
\cite{pinto17} proposed to model the dynamics between the adversary and the deep neural policy as a zero-sum game \citep{littman94} where the goal of the adversary is to minimize expected cumulative rewards of the deep reinforcement learning policy. 
\[
R^{\textrm{agent}} = \mathbb{E}_{s_0 \sim \rho ; a^{\textrm{agent}} \sim \pi^{\textrm{agent}} ; a^{\textrm{adv}} \sim \pi^{\textrm{adv}} }  [\sum_{t=0}^{T-1} r^{\textrm{agent}}(s,a^{\textrm{agent}}, a^{\textrm{adv}})]
\]
\begin{figure}[t]
\footnotesize
\centering
\stackunder[0pt]{\includegraphics[scale=0.26]{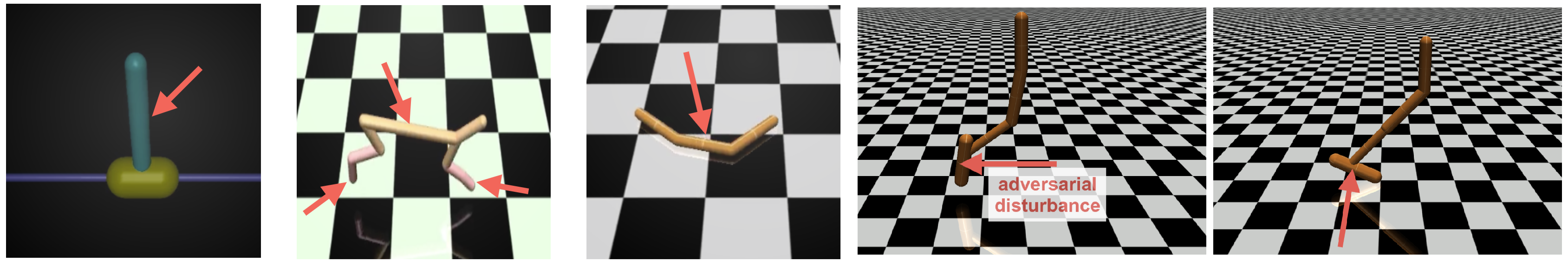}}{}
\vskip -0.1in
\caption{Robust adversarial reinforcement learning proposed in \citep{pinto17}. This paper proposes the zero-sum game to model the relationship between the agent and the adversary while focusing on introducing disturbances to the environment dynamics. Here the empirical studies are conducted in the MuJoCo environment.}
\label{dataaug}
\end{figure}
Here the adversarial policy is represented by $\pi^{\textrm{adv}}$, the policy of the agent represented by $\pi^{\textrm{agent}}$, and the rewards received by the agent represented by $r^{\textrm{agent}}$. 
The Nash equilibrium of the optimal rewards for this zero-sum game is
\[
R^{\textrm{agent}^*} 
= \min_{\pi^{\textrm{adv}} } \max_{\pi^{\textrm{agent}}} R^{\textrm{agent}}(\pi^{\textrm{agent}}, \pi^{\textrm{adv}}) 
=  \max_{\pi^{\textrm{agent}}} \min_{\pi^{\textrm{adv}} } R^{\textrm{agent}}(\pi^{\textrm{agent}}, \pi^{\textrm{adv}}) 
\]
This study is a clear example of transition probability perturbation to achieve generalization $\mathcal{G}_\mathcal{P}$ in Definition \ref{def:transitionperturbing} of Section \ref{def}.
\cite{glaeve19} approached this problem with an adversary model which is restricted to take natural actions in the MDP instead of modifying the observations with $\ell_p$-norm bounded perturbations. The authors model this dynamic as a zero-sum Markov game and solve it via self play Proximal Policy Optimization (PPO).
Some recent studies, proposed to model the interaction between the adversary and the deep reinforcement learning policy as a state-adversarial MDP, and claimed that their proposed algorithm State Adversarial Double Deep Q-Network (SA-DDQN) learns theoretically certified robust policies against natural noise and perturbations. In particular, these certified adversarial training techniques aim to add a regularizer term to the temporal difference loss in deep $Q$-learning 
\[
\mathcal{H}(r_i + \gamma \max_{a} \hat{Q}_{\hat\theta}(s_i, a;\theta) - Q_\theta(s_i,a_i;\theta)) + \kappa \mathcal{R}(\theta)
\]
where $\mathcal{H}$ is the Huber loss, $\hat{Q}$ refers to the target network and $\kappa$ is to adjust the level of regularization for convergence. The regularizer term can vary for different certified adversarial training techniques yet the baseline technique uses $\mathcal{R}(\theta)$
\begin{align*}
\max \{ \max_{\hat{s} \in B(s)}  \max_{a \neq \argmax_{a'}  Q(s,a')}  Q_{\theta}(\hat{s},a)  
 - Q_{\theta}(\hat{s},\argmax_{a'} Q(s,a')), -c \}.
\end{align*}
where $B(s)$ is an $\ell_p$-norm ball of radius $\epsilon$.
While these certified adversarial training techniques drew some attention from the community, more recently manifold concerns have been raised on the robustness of theoretically certified adversarially trained deep reinforcement learning policies \citep{korkmazmap21, korkmazuai,korkmaz22,korkmazneurips24}. In these studies, the authors argue that adversarially trained (i.e. certified robust) deep reinforcement learning policies learn inaccurate state-action value functions and non-robust features from the environment.
More importantly, recently it has been shown that certified robust deep reinforcement learning policies have worse generalization capabilities compared to vanilla trained reinforcement learning policies in high dimensional state space MDPs \citep{korkmaz2023aaai}.
While this study provides a contradistinction between adversarial and natural directions that are intrinsic to the MDP, it further demonstrates that the certified adversarial training techniques block generalization capabilities of standard deep reinforcement learning policies.
Furthermore note that this study is also a clear example of a state observation perturbation generalization testing method $\mathcal{G}_S^S$ in Definition \ref{def:gentesting} in Section \ref{def}.
For a more comprehensive view on generalization and robustness see \cite{korkmazthesis24}.

It is important to observe that the methods that focuses on improving generalization, i.e. robust training, described in this section rarely employ the different generalization testing methods proposed by other work. Thus, focusing narrowly on one aspect of generalization with one dimensional improvements in actuality decreases generalization on another aspect, as has been shown in the case of adversarial training \citep{korkmaz2023aaai}. Therefore we again emphasize the need to understand the significance of a concrete definition of generalization, and a unified baseline to precisely measure it.

\begin{figure}[t]
\footnotesize
\centering
\stackunder[0pt]{\includegraphics[scale=0.136]{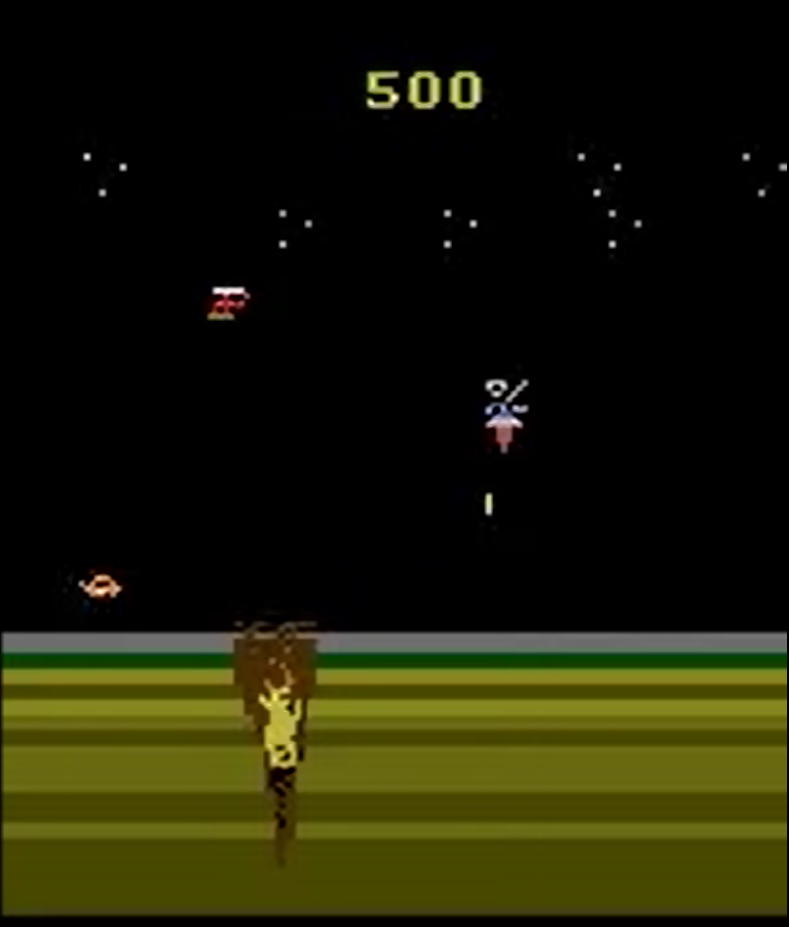}}{}
\hskip 0.1pt
\stackunder[0pt]{\includegraphics[scale=0.068]{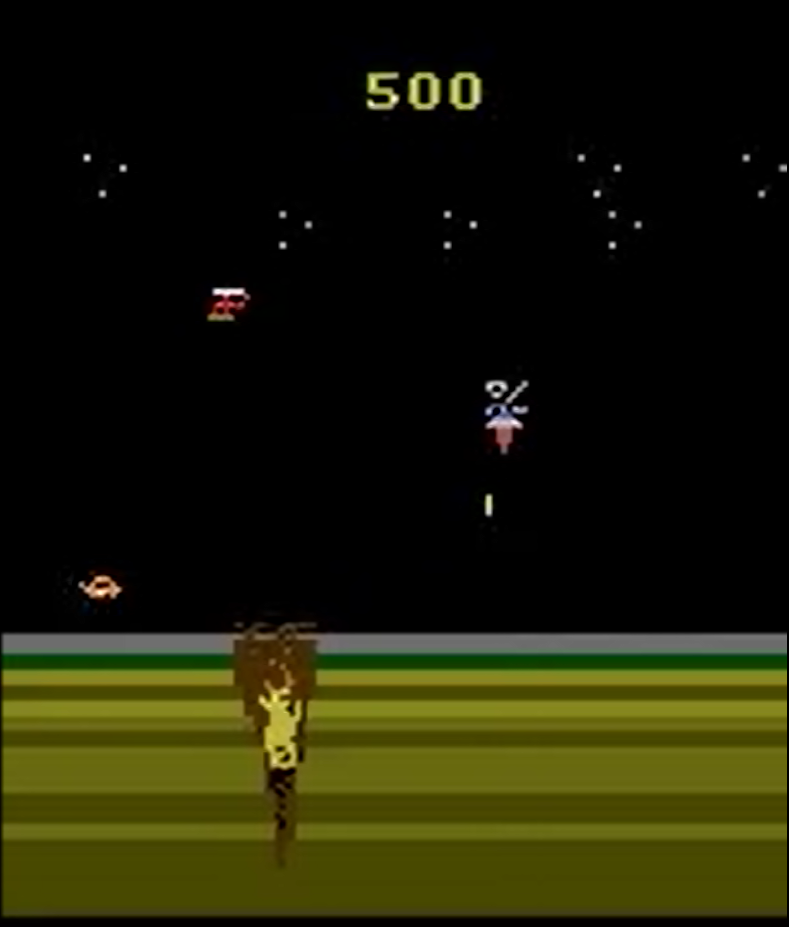}}{}
\stackunder[0pt]{\includegraphics[scale=0.068]{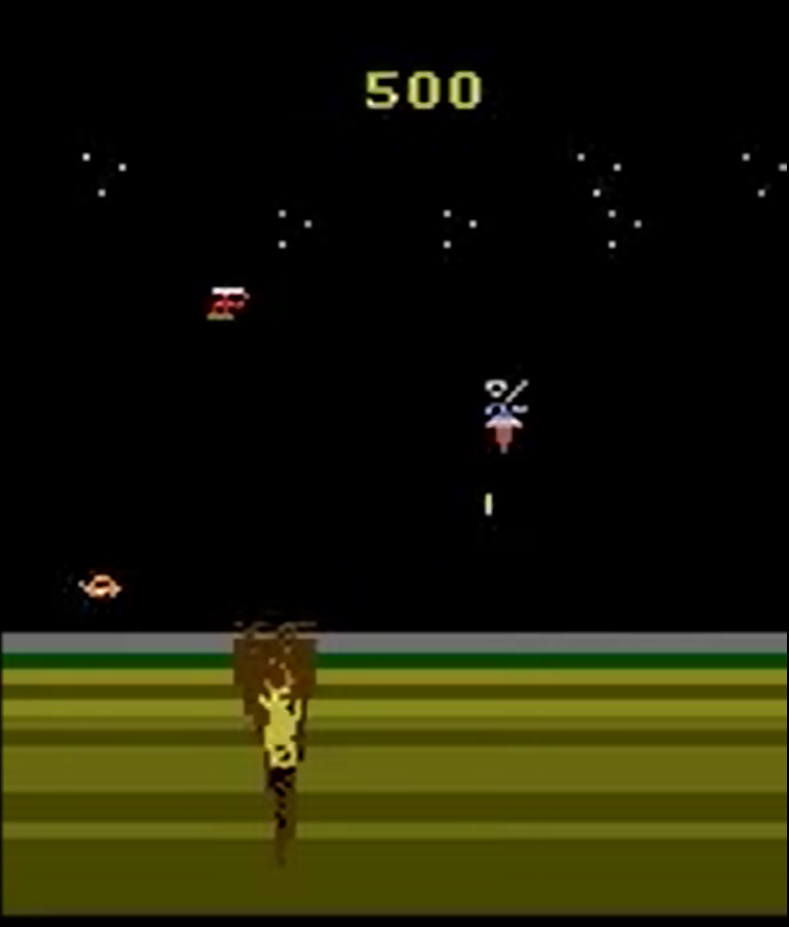}}{}
\stackunder[0pt]{\includegraphics[scale=0.068]{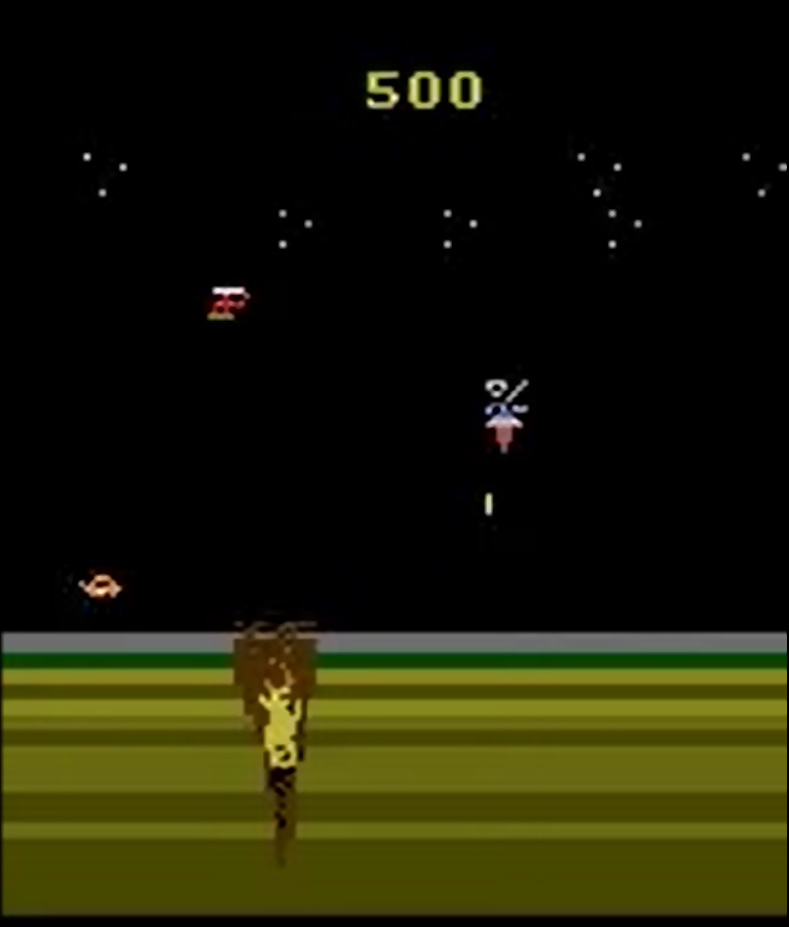}}{}
\stackunder[0pt]{\includegraphics[scale=0.068]{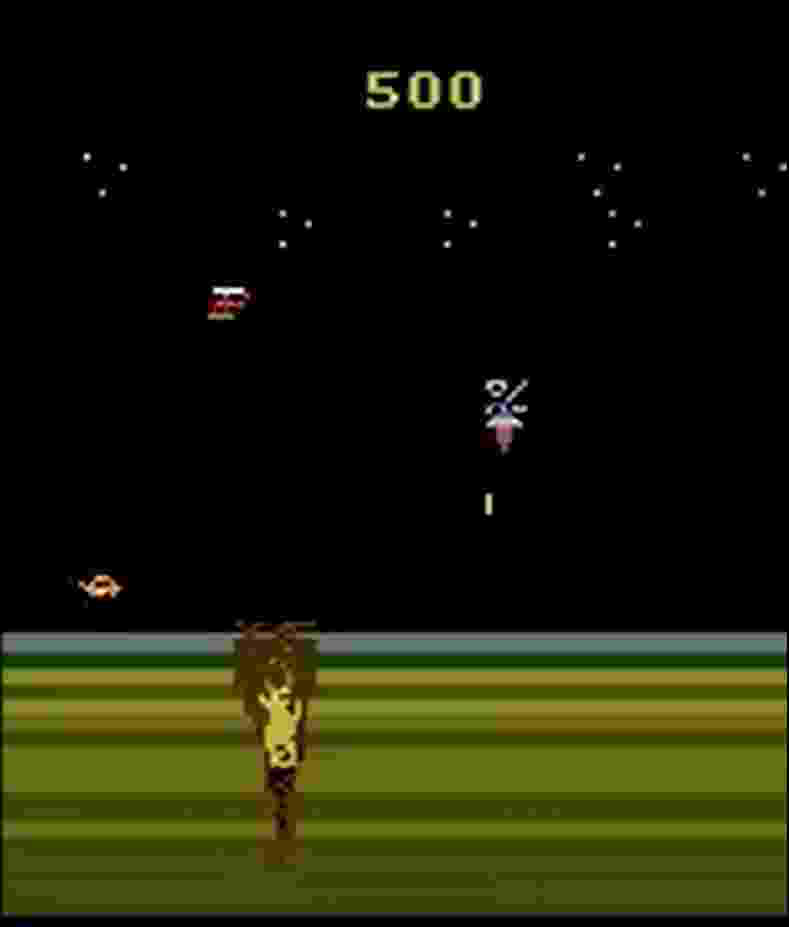}}{}
\stackunder[0pt]{\includegraphics[scale=0.068]{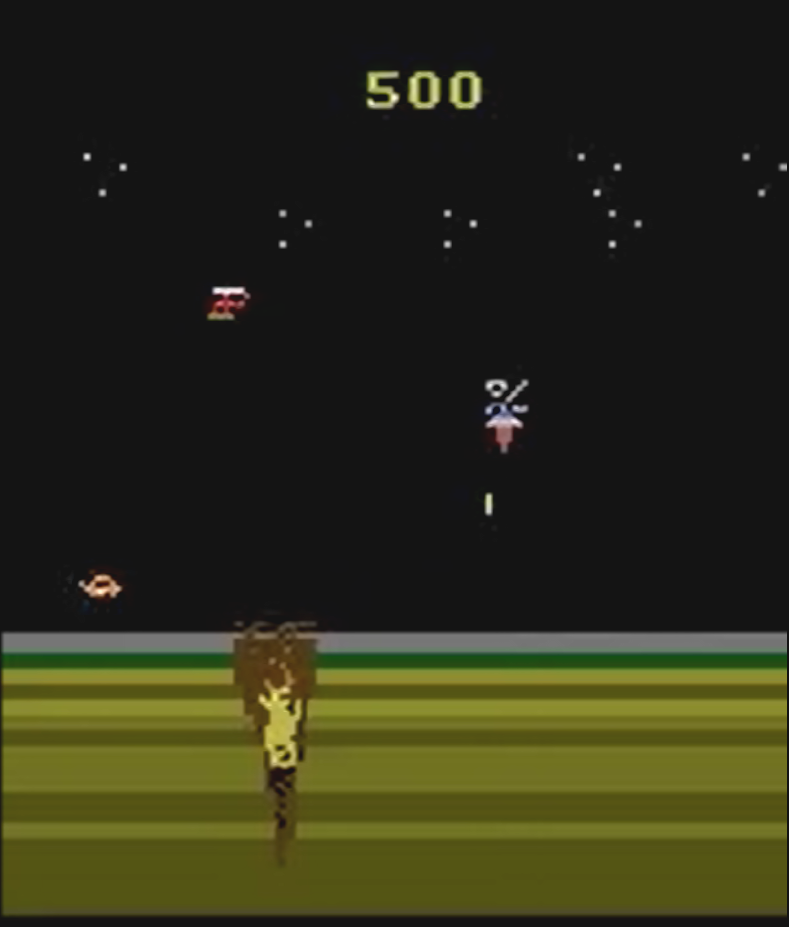}}{} \\
\stackunder[6pt]{\includegraphics[scale=0.1465]{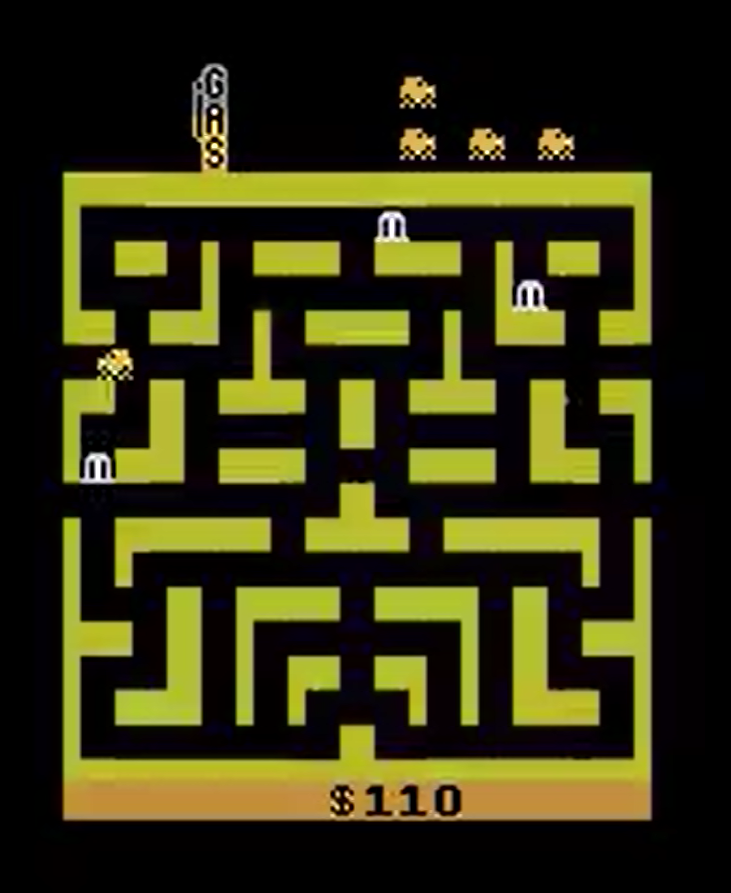}}{\scriptsize{Base State}}
\hskip 0.1pt
\stackunder[6pt]{\includegraphics[scale=0.074]{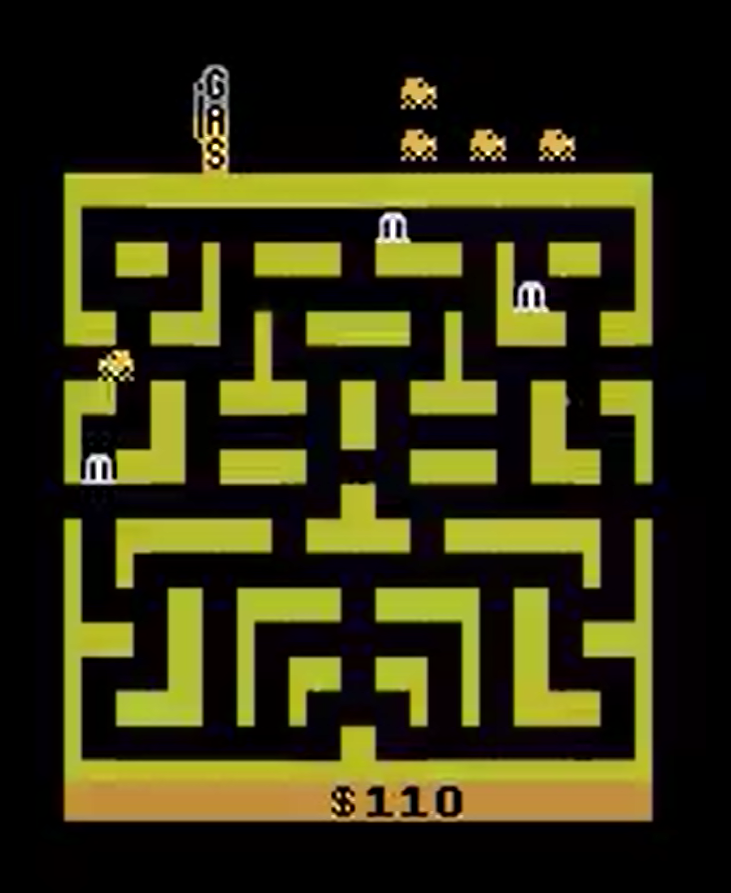}}{\scriptsize{Shift}}
\stackunder[6pt]{\includegraphics[scale=0.074]{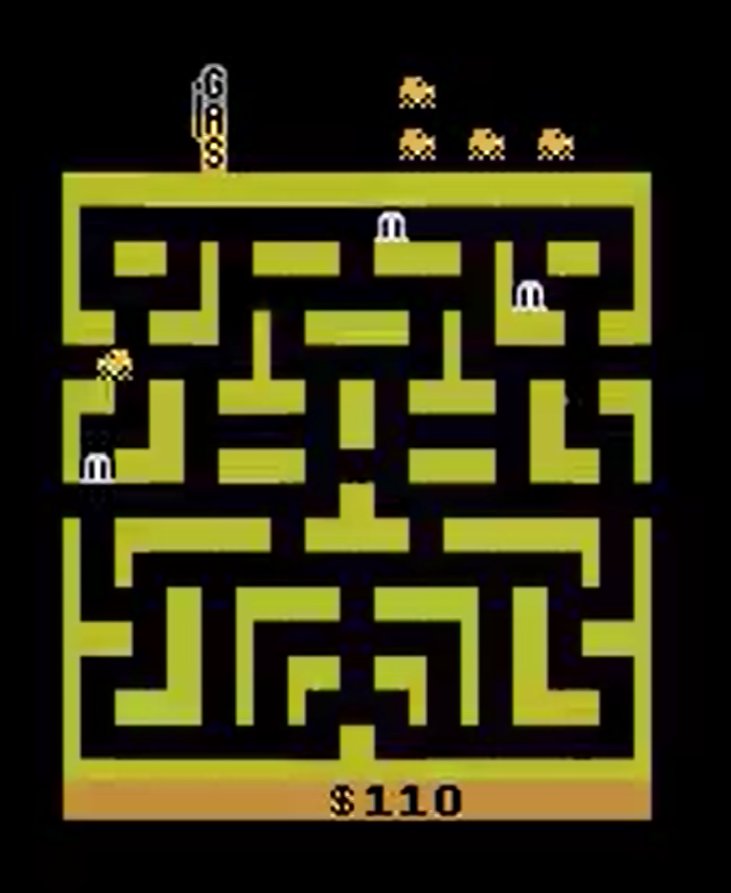}}{\scriptsize{PT}}
\stackunder[6pt]{\includegraphics[scale=0.074]{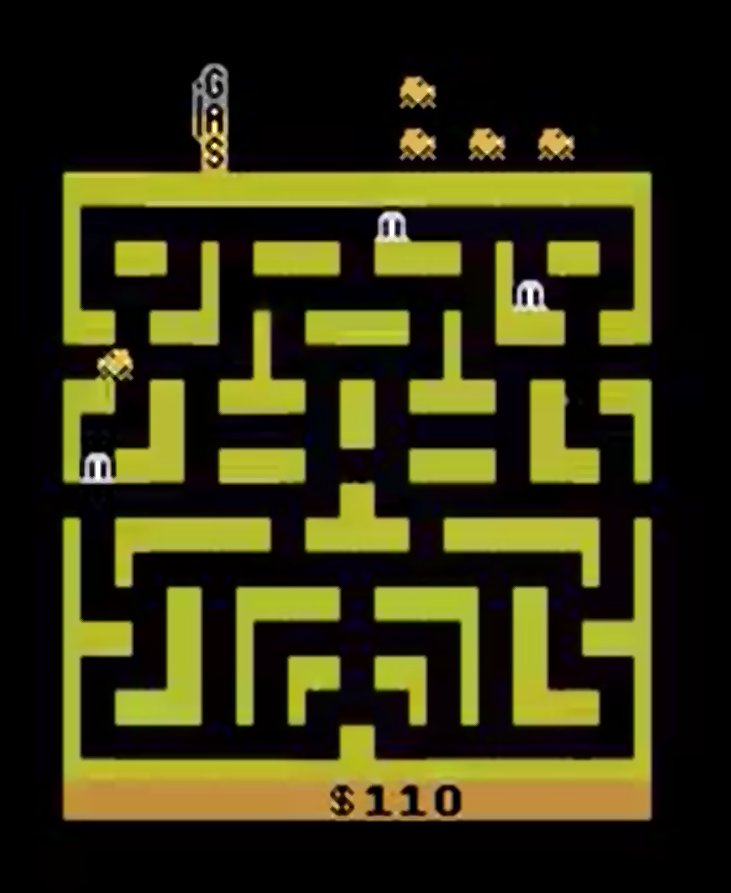}}{\scriptsize{Blur}}
\stackunder[6pt]{\includegraphics[scale=0.074]{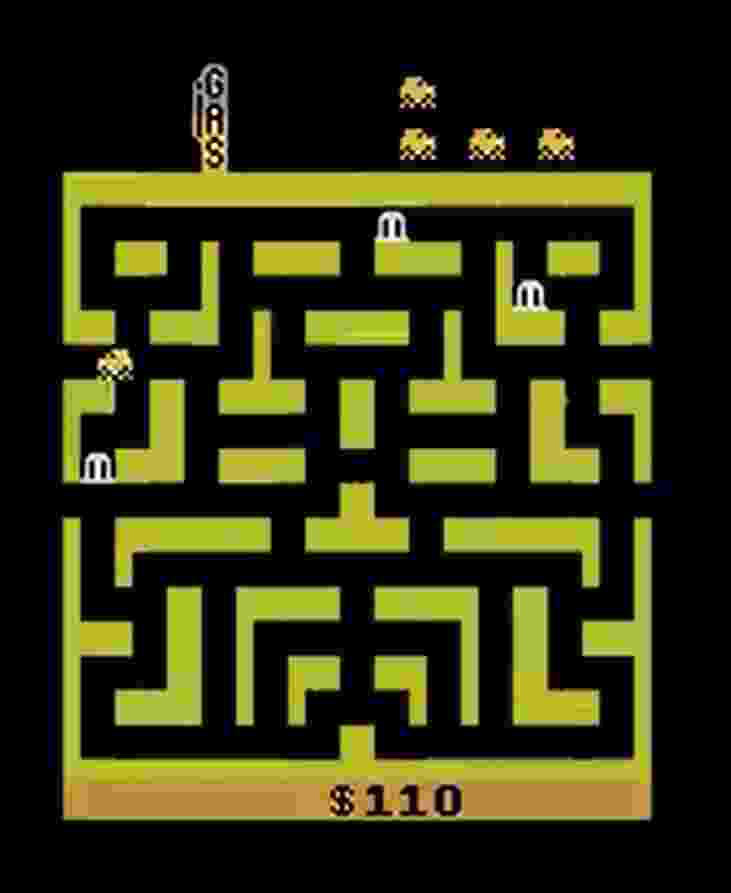}}{\scriptsize{DCT}}
\stackunder[6pt]{\includegraphics[scale=0.074]{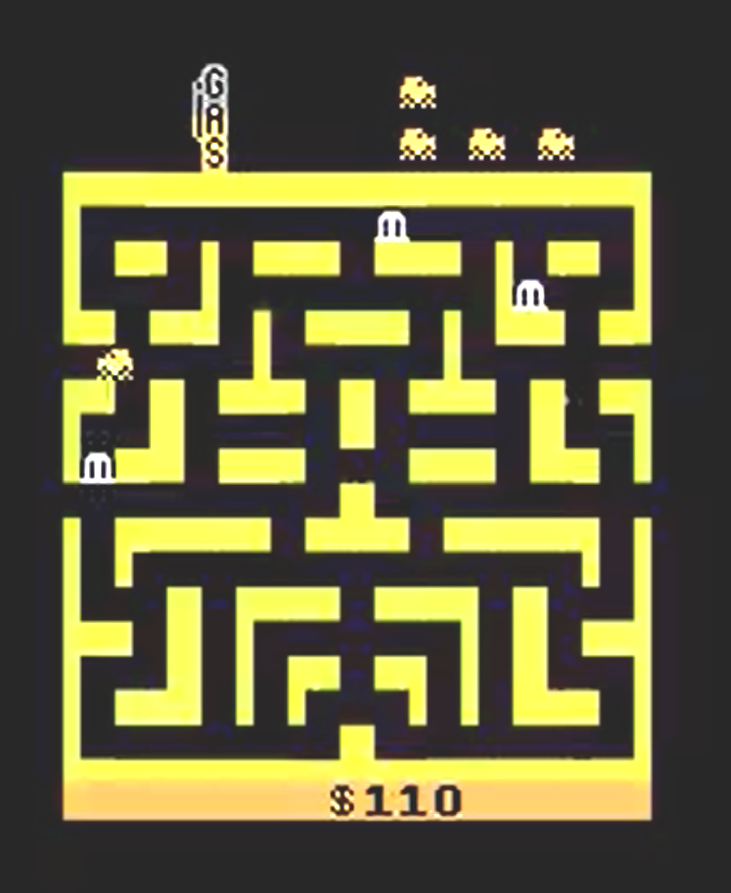}}{\scriptsize{B\&C}} \\
\vskip -0.1in
\caption{State transformation generalization under adversarial perspective in the Arcade Learning Environment \citep{korkmaz2023aaai}. Note that under the adversarial influence direction of research, the state transformation generalization is constrained by the imperceptibility of the transformations. Columns: base frame, shifting, perspective transformation, blurring, discrete cosine transform artifacts, brightness and contrast. Up: JamesBond. Down: BankHeist. }
\label{visual}
\vskip -0.1in
\end{figure}

\section{Meta-Reinforcement Learning and Meta Gradients}
\label{metarlsec}
A quite recent line of research directs its research efforts to discovering reinforcement learning algorithms automatically, without explicitly designing them, via meta-gradients \citep{oh20, xu20}. This line of study targets learning the "learning algorithm" by only interacting with a set of environments as a meta-learning problem. In particular, 
\[
\eta^* = \argmax_\eta \mathbb{E}_{\varepsilon \sim \rho(\varepsilon)} \mathbb{E}_{\theta_0 \sim \rho(\theta_0)}
[\mathbb{E}_{\theta_N} [\sum_{t=0}^\infty  \gamma^t r_t ]    ]
\]
here the optimal update rule is parametrized by $\eta$, for a distribution on environments $\rho(\varepsilon)$ and initial policy parameters $\rho(\theta_0)$ where $\mathbb{E}_{\theta_N} [\sum_{t=0}^\infty  \gamma^t r_t ]$ is the expected return for the end of the lifetime of the agent.
The objective of meta-reinforcement learning is to be able to build agents that can learn \textit{how to learn} over time, thus allowing these policies to adapt to a changing environment or even any other changing conditions of the MDP.

Quite recently, a significant line of research has been conducted to achieve this objective, particularly \cite{oh20} proposes to discover update rules for reinforcement learning. 
This line of work also falls under the algorithmic generalization $\mathcal{G}_\mathbb{A}$ in Definition \ref{def:algorithmicgeneralization} defined in Section \ref{def}. Following this work \cite{xu20} proposed a joint meta-learning framework to learn what the policy should predict and how these predictions should be used in updating the policy.
Recently, \cite{kirsch22} proposes to use symmetry information in discovering reinforcement learning algorithms and discusses meta-generalization.
There is also some work on enabling reinforcement learning algorithms to discover temporal abstractions \citep{vivek21}. In particular, temporal abstraction refers to the ability of the policy to abstract a sequence of actions to achieve certain sub-tasks.
As it is promised within this subfield, meta-reinforcement learning is considered to be a research direction that could enable us to build deep reinforcement learning policies that can generalize to different environments, to changing environments over time, or even to different tasks.

\begin{figure}[t]
\footnotesize
\centering
\stackunder[0pt]{\includegraphics[scale=0.4]{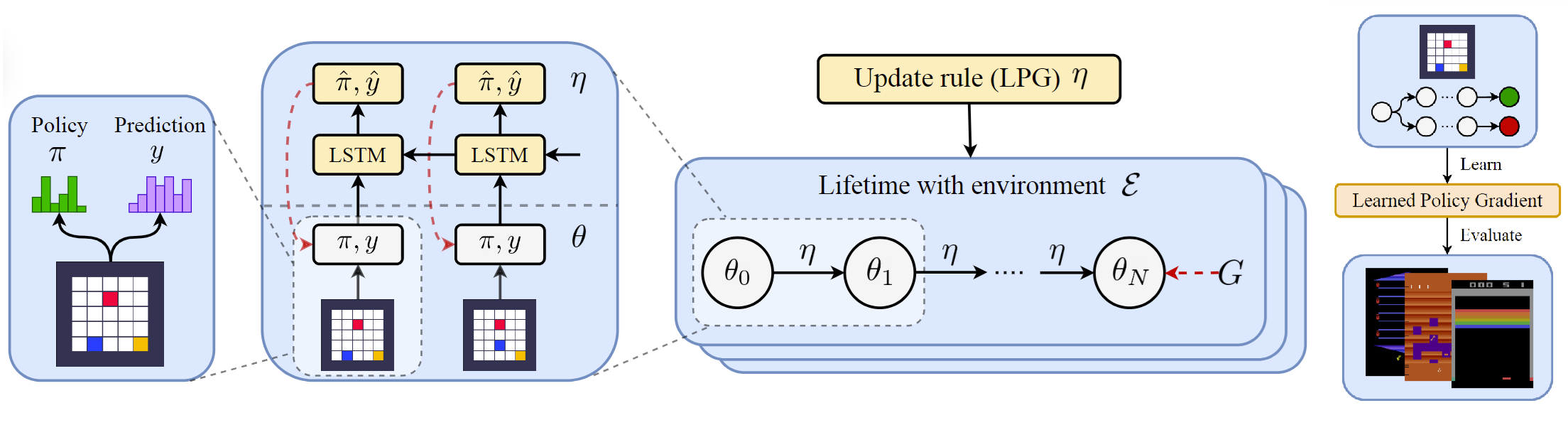}}{}
\vskip -0.1in
\caption{Meta training of the learned policy gradient that have been described in \citep{oh20}. Right: The learned policy gradient algorithm that has been trained in toy examples can generalize to more complex environment such as the Arcade Learning Environment.}
\label{metarl}
\vskip -0.1in
\end{figure}

\section{Transfer in Reinforcement Learning}
\label{transfer}

Transfer in reinforcement learning is a subfield heavily discussed in certain applications of reinforcement learning algorithms, e.g. robotics. In current robotics research there is not a safe way of training a reinforcement learning agent by letting the robot explore in real life. Hence, the way to overcome this is to train policies in a simulated environment, and install the trained policies in the actual application setting. The fact that the simulation environment and the installation environment are not identical is one of the main problems for reinforcement learning application research. This is referred to as the \textit{sim-to-real gap}.

Another subfield in reinforcement learning research focusing on obtaining generalizable policies investigates this concept through \textit{transfer in reinforcement learning}. The consideration in this line of research is to build policies that are trained for a particular task with limited data and to try to make these policies perform well on slightly different tasks. An initial discussion on this starts with \cite{stone07} to obtain policies initially trained in a source task and transferred to a target task in a more sample efficient way. Later, \cite{andrea18} proposes to transfer value functions that are based on learning a prior distribution over optimal value functions from a source task. However, this study is conducted in simple environments with low dimensional state spaces.
\cite{barr17} considers the reward transformation setting $\mathcal{G}_R$ in Definition \ref{def:rewardsperturbing} from Section \ref{def}. In particular, the authors consider a policy transfer between a specific task with a reward function $r(s,a)$ and a different task with reward function $r'(s,a)$. The goal of the study is to decouple the state representations from the task. In the setting of state transformation for generalization $\mathcal{G}_S$ in Definition \ref{def:stateperturbing} \cite{gamrian19} focuses on state-wise differences between source and target task. In particular, the authors use unaligned generative adversarial networks to create target task states from source task states.
In the setting of policy transformation for generalization $\mathcal{G}_\pi$ in Definition \ref{def:policyperturbing} \cite{jain20} focuses on zero-shot generalization to a newly introduced action set to increase adaptability.
While transfer learning is a promising research direction for reinforcement learning, the studies in this subfield still remain oriented only towards reinforcement learning applications, and thus the main focus on applications centered on this subfield provides a non-unified progress in research due to the lack of an established baseline in which the proposed claims and algorithms can be consistently compared.

\begin{figure}[t]
\footnotesize
\centering
\stackunder[0pt]{\includegraphics[scale=0.24]{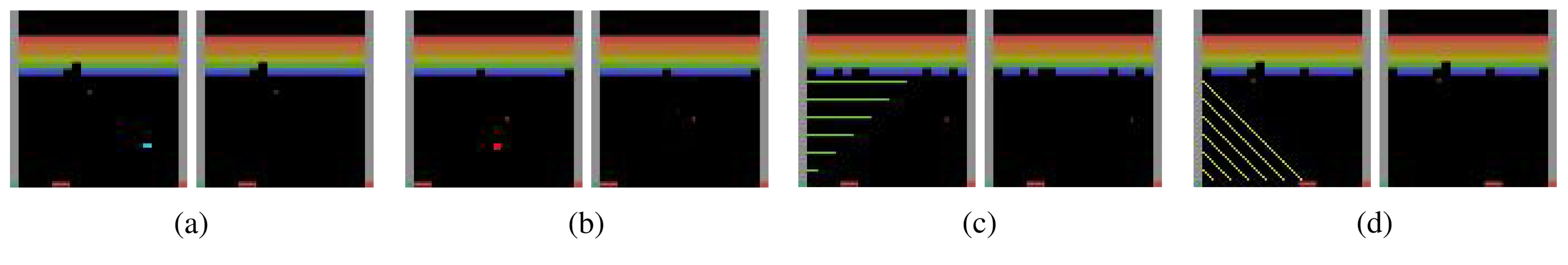}}{}
\vskip -0.1in
\caption{Transfer in reinforcement learning as has been described in \citep{gamrian19} that falls under the generalization through observation category explained in Definition \ref{def:stateperturbing}. The frames are taken from Breakout game in the Arcade Learning Environment. The left frames represent the target task and the right frames represents the source tasks generated via generative adversarial networks.}
\label{dataaug}
\vskip -0.1in
\end{figure}

\section{Lifelong Reinforcement Learning}

\textit{Lifelong learning} is a subfield closely related to transfer learning that has recently drawn attention from the reinforcement learning community. Lifelong learning aims to build policies that can sequentially solve different tasks by being able to transfer knowledge between tasks. On this line of research, \cite{leca21} provide an algorithm for value-based transfer in the Lipschitz continuous task space with theoretical contributions for lifelong learning goals. In the setting of action transformation for generalization $\mathcal{G}_\pi$ in Definition \ref{def:policyperturbing} \cite{chandak20} focuses on temporally varying (e.g. variations between source task and target task) the action set in lifelong learning. In lifelong reinforcement learning some studies focus on different exploration strategies. In particular, \cite{garcia19} models the exploration strategy problem for lifelong learning as another MDP, and the study uses a separate reinforcement learning agent to find an optimal exploration method for the initial lifelong learning agent. The lack of benchmarks limits the progress of lifelong reinforcement learning research by restricting the direct comparison between proposed algorithms or methods. However, quite recent work proposed a new training environment benchmark based on robotics applications for lifelong learning to overcome this issue \citep{zajac21}\footnote{The state dimension for this benchmark is 12. Hence, the state space is low dimensional.}.

\section{Inverse Reinforcement Learning}

\textit{Inverse reinforcement learning} focuses on learning a functioning policy in the absence of a reward function. Since the real reward function is inaccessible in this setting and the reward function needs to be learnt from observing an expert completing the given task, the inverse reinforcement learning setting falls under the reward transformation for generalization setting $\mathcal{G}_R$ defined in Definition \ref{def:rewardsperturbing} in Section \ref{def}. The initial work that introduced inverse reinforcement learning was proposed by \cite{ng00} demonstrating that multiple different reward functions can be constructed for an observed optimal policy. The authors of this initial study achieve this objective via linear programming,
\begin{align*}
\textrm{max} \sum_{s\in S_{\rho}} & \min_{a \in A} \{ p(\mathbb{E}_{s' \sim \mathcal{P}(s,a_1|\cdot)} \mathcal{V}^\pi(s') -\mathbb{E}_{s' \sim \mathcal{P}(s,a|\cdot)} \mathcal{V}^\pi(s')) \} \\  \nonumber
& \textrm{s.t.} \:\: |\alpha_i|\leq 1 \:,\: i=1,2, \dots, d
\end{align*}
where $p(x)=x$ if $x \geq 0$, $p(x)=2x$ otherwise and $\mathcal{V}^\pi = \alpha_1\mathcal{V}^\pi_1 + \alpha_2\mathcal{V}^\pi_2 + \dots + \alpha_d\mathcal{V}^\pi_d$.
In this line of work, there has been recent progress that achieved learning functioning policies in high-dimensional state observation MDPs \citep{garg21}. The study achieves this by learning a soft $Q$-function from observing expert demonstrations, and the study further argues that it is possible to recover rewards from the learnt soft state-action value function.

\section{Conclusion}

In this paper we tried to answer the following questions: 
\textit{(i) What are the explicit problems limiting reinforcement learning algorithms from obtaining high-performing policies that can generalize to complex environments? 
(ii) How can we unify and categorize the concept of generalization in deep reinforcement learning considering many subfields under reinforcement learning at their core focus on the same objective?
(iii) What are the similarities and differences of these different techniques proposed by different subfields of reinforcement learning research to build reinforcement learning policies that can robustly generalize?} 
To answer these questions first we introduce a theoretical analysis and mathematical framework to unify and categorize the concept of generalization in deep reinforcement learning. Then we explain the connection and the significance of exploration in overfitting to a learning environment, and explain the manifold causes of overestimation bias in reinforcement learning. 
Starting from all the different regularization techniques in either state representations or in learnt value functions from worst-case to average-case, we provide a current layout of the wide range of reinforcement learning subfields that are essentially working towards the same objective, i.e. generalizable deep reinforcement learning policies.
Finally, we provided a discussion for each category on the drawbacks and advantages of these algorithms. We believe our study can provide a compact unifying formalization on recent reinforcement learning generalization research.
We believe our theoretical framework can guide current and future research to build deep reinforcement learning agents that can robustly generalize to complex environments.

\vskip 0.2in
\bibliographystyle{apalike}
\bibliography{example_paper.bib}

\end{document}